\documentclass[pre,twocolumn,a4paper,showpacs]{revtex4-1}
\usepackage{amssymb}
\usepackage{wasysym}
\usepackage{graphicx}
\usepackage{epsfig}
\usepackage{subfigure}
\begin{document}

\title{Awareness improves problem-solving performance}

\author{Jos\'e F. Fontanari}

\affiliation{Instituto de F\'{\i}sica de S\~ao Carlos,
  Universidade de S\~ao Paulo,
  Caixa Postal 369, 13560-970 S\~ao Carlos, S\~ao Paulo, Brazil}

\begin{abstract}
The brain's  self-monitoring of activities, including internal activities -- a functionality that we  refer to as  awareness -- has been suggested as   a key element of consciousness. Here we investigate whether the presence of  an inner-eye-like process (monitor) that supervises the activities of a number of subsystems (operative agents) engaged in the solution of a problem can improve the problem-solving efficiency of the system. The problem  is to find  the global maximum of a NK fitness landscape and the performance is measured by the time required to find that  maximum. The operative agents explore blindly the fitness landscape and the monitor provides them with feedback on the quality (fitness) of the proposed solutions. This feedback is then used by the operative agents to bias their searches towards the fittest regions of the landscape.
We find that a weak feedback between the monitor and the operative agents improves the performance of the system, regardless of the difficulty of the problem, which is gauged by the number of local maxima in the landscape. For easy problems (i.e., landscapes without local maxima), the performance improves monotonically as the feedback strength  increases, but for difficult problems, there is an optimal 
value of the feedback strength beyond which   the system  performance degrades very rapidly.
\end{abstract}

\maketitle

\section{Introduction}

What is consciousness for? From a biological perspective, an auspicious answer to this mind-opening question  (see \cite{Blackmore_03} for a thorough discussion of the theories of consciousness) views consciousness as a source of information about brain states --  a  brain's schematic description of those states -- and suggests that the evolutionary usefulness of such inner eye  is to provide human beings  with an effective tool for doing natural psychology, i.e., for imagining what might be happening inside another person's head \cite{Humphrey_99}. Hence, the conception of other people as beings with minds originates from  the way each individual sees himself and,  in that sense, solely extraordinarily  social creatures, probably humans only, would  evolve consciousness as a response  to  the pressures to handle  interpersonal relationships  \cite{Humphrey_99}. There is an alternative, equally attractive, possibility  that  we   may  first   unconsciously    suppose   other  consciousness,  and  then  infer  our  own  by  generalization \cite{Jaynes_76}.  We note that the hypothesis that consciousness is closely related to social ability has been suggested in many forms by many authors (see, e.g., \cite{Frith_95,Perlovsky_06,Carruthers_09,Baumeister_10}), but the original insight that  consciousness and  cognition are  products of social behaviors   probably dates back to  Vygostsky in the 1930s \cite{Vygotsky_86}.  

This approach, however, is not very  helpful to the engineer who  wants to build a conscious machine.  Fortunately, the recently proposed  attention schema theory of consciousness \cite{Graziano_11,Graziano_13} offers some hope to our engineer by positing  that awareness  is simply  a schematic model of one's state of attention, i.e., awareness is an internal model of attention.  (The intimate connection between awareness and  consciousness  is  expressed best  by the view that consciousness is simply the awareness  of  what we have done or said, reflected back to us \cite{Jaynes_76}.) Building a functioning attention schema is a feasible  software project today, which could then be coupled to the existing perceptual schemas \cite{Murphy_00} to create a conscious machine.  As before,  the selective value of such internal model stems  from the possibility of attributing the same model to other people, i.e, of doing natural psychology  \cite{Graziano_13}. 

Internal models or inner eyes  keep track of processes that, within an evolutionary perspective, are  useful to monitor and provide feedback  to (or report on)  those very same processes. This feedback can be thought of as the mechanism by which   `mind'  influences matter \cite{Graziano_13}.  Here we show that the inner monitoring can be useful in a more general problem-solving scenario. (The word awareness in the title of this paper is used with the meaning of  inner monitoring.)   In particular,   we consider a number $L$ of subsystems or operative agents that search randomly for the solution of a problem, viz. finding the global maximum of a rugged fitness  landscape (see Section \ref{sec:NK}), and a single monitor that tracks the quality of the solution found by each agent (i.e., its fitness) and records the best solution at  each time. The feedback to the operative agents occurs with frequency $p \leq 1$,  i.e., on the average each agent receives feedback from the monitor $p \times \Delta t$ times during the time interval $\Delta t$. The feedback consists of displaying the best solution among all agents at that time, so the operative agents can copy small pieces of that solution (see Section \ref{sec:model} for details).

The performance of the system composed of  $L$ operative agents and a monitor is measured, essentially, by the time it takes to find the global maximum of the fitness landscape. (Since we may want to compare performances for different values of $L$, the actual performance measure must be  properly scaled by $L$, as discussed in Section \ref{sec:model}) The relevant comparison is between the case $p=0$ where the monitor has no effect on the operation of the system (a scenario akin to the doctrine of epiphenomenalism \cite{Blackmore_03}), and the case $p > 0$ where the system receives feedback from the monitor.   If the speed to solve problems has a survival value to the individuals and  if that speed increases in the presence of feedback from the monitor, then one may argue for the plausibility of the  evolution, as well as for the commonplaceness, of  such inner-eyes-like processes  in the brain.

We find that the performance of the system for small values of the feedback frequency or  strength $p$, which is likely the most realistic scenario,  is  superior  to the performance in absence of feedback, regardless of the difficulty of the task and of the size of the system. This
finding lends support to the inner-eye scenario for brain processes. In the case of easy tasks (i.e., landscapes without local maxima), the performance always improves with increasing  $p$, but for rugged landscapes the situation is more complicated: there exists an optimal value of $p$, which depends both on the  complexity of the task and on the system size, beyond which the system performance deteriorates abruptly.

The rest of this paper is organized as follows. Since the  tasks of varying complexity presented to the problem-solving system are finding the global maxima  of  rugged fitness landscapes generated by the NK model, in Section \ref{sec:NK} we offer  an outline of that classic model \cite{Kauffman_87}. The problem-solving system is then described  in great detail in Section \ref{sec:model}. 
We explore the space of parameters of the problem-solving system as well as of the NK model in   Section \ref{sec:res}, where we present and analyze the results of our simulations.     Finally, Section \ref{sec:disc} is reserved to our concluding remarks.

\section{Task}\label{sec:NK}

The task posed to a system of $L$  agents is  to find the unique global maximum of  a fitness landscape generated using   the
NK model  \cite{Kauffman_87}.  For our purposes, the advantage of using the  NK model is that  it allows the  tuning  of the ruggedness of the landscape -- and hence of the difficulty of the task -- by changing the integer parameters  $N$ and $K$. More specifically,  the NK landscape is defined in the space of binary strings $\mathbf{x} = \left ( x_1, \ldots, x_N \right ) $ with $x_i = 0,1$  and so the parameter $N$ determines the size of the state space, given by $2^N$.  
For each bit string $\mathbf{x}$ is assigned a distinct real-valued fitness value $ \Phi \left ( \mathbf{x} \right ) \in \left [ 0, 1 \right ]$ which  is an average  of the contributions from each  element $i$  of the string, i.e.,
\begin{equation}
\Phi \left ( \mathbf{x}  \right ) = \frac{1}{N} \sum_{i=1}^N \phi_i \left (  \mathbf{x}  \right ) ,
\end{equation}
where $ \phi_i$ is the contribution of element $i$ to the  fitness of string $ \mathbf{x} $. It is assumed that $ \phi_i$ depends on the state
$x_i$  as well as on the states of the $K$ right neighbors of $i$, i.e., $\phi_i = \phi_i \left ( x_i, x_{i+1}, \ldots, x_{i+K} \right )$ with the arithmetic in the subscripts done modulo $N$. The  parameter  $K =0, \ldots, N-1$ is called the degree of epistasis and  determines the ruggedness of  the landscape for fixed $N$.
The functions $\phi_i$ are $N$ distinct real-valued functions on $\left \{ 0,1 \right \}^{K+1}$ and, as usual,  we assign to each $ \phi_i$ a uniformly distributed random number  in the unit interval so that   $\Phi \in \left ( 0, 1 \right )$ has a unique global maximum \cite{Kauffman_87}.

The increase of the parameter $K$ from $0$ to $N-1$  decreases the correlation between the fitness of neighboring strings
(i.e.,  strings that differ at a single bit) in the state space. In particular, the local fitness correlation is given by 
$ corr \left ( \mathbf{x}, \tilde{\mathbf{x}}_i \right ) = 1 - \left ( K+1 \right )/N $  where $\tilde{\mathbf{x}}_i$ is the string  $\mathbf{x} $ with
bit $i$ flipped. Hence 
for $K=N-1$ the fitness values are  uncorrelated and the NK model reduces to the Random Energy model \cite{Derrida_81,Saakian_09}.  
Finding   the global maximum of the NK model for $K>0$ is an NP-complete problem \cite{Solow_00}, which  means that the time required to solve {\it all}  realizations of that landscape using any currently known deterministic algorithm increases exponentially fast with the length $N$ of the strings.  However, for $K=0$ the (smooth) landscape has a single maximum that is easily located by picking for each string element $i$ the state $x_i = 0$ if  $\phi_i \left ( 0 \right ) >  \phi_i \left ( 1 \right )$ or the state  $x_i = 1$, otherwise. On the average,  the number of local maxima increases with increasing  $K$.  This number  can be associated with the difficulty of the task provided the search heuristic explores the local correlations of fitness values 
to locate the global maximum of the fitness landscape, which is the case  of the  search heuristic used in our simulations.

Since the fitness values  $\Phi \left ( \mathbf{x} \right )$ are random, the number of local maxima    varies considerably between landscapes characterized by the same values of $N$ and $K>0$, which makes the performance of any search  heuristic based on the  local correlations of the fitness landscape strongly  dependent  on the particular realization of the landscape. Hence  we evaluate the system performance  in a sample of 100 distinct realizations of the NK fitness landscape for fixed $N$ and $K$.
In particular,  we fix the string length  to $N=16$  and allow the degree of epistasis to take on the values $K=0$, $1$, $3$ and $5$. In addition,  in order to study landscapes with different state space sizes but the same correlation between the fitness of neighboring states
we consider also strings of length $N=12$ and $N=20$. 

\begin{table}
\caption{Statistics of the number of maxima in the sample of 100 NK-fitness landscapes  used in the computational experiments.}
\centering
\begin{tabular}{c c c c c}
\hline
N & \hspace{.5 cm} K  & \hspace{1cm} mean  \hspace{1cm} &  min  & max \\ [0.5ex]
\hline
16 & ~~~~~0&  1   &  1 & 1  \\
16 & ~~~~~1 & 8.4  & 1 & 32  \\
16 & ~~~~~3 & 84.7  & 26 & 161 \\
16 &~~~~~5 & 292.1  & 235 & 354 \\  
12 & ~~~~~2 & 13.1  & 4 & 29 \\
20 & ~~~~~4 & 633.0  & 403 & 981 \\ [1ex]
\hline
\end{tabular}
\label{table:1}
\end{table}
%

Table  \ref{table:1} shows the mean number of maxima,   as well as two extreme statistics, namely,  the minimum and  the maximum number of maxima, in the sample of 100 landscapes used in the computational experiments. Although the landscapes $(N=12,K=2)$, $(N=16,K=3)$ and $(N=20,K=4)$ exhibit the same local  fitness correlation, viz. $ corr \left ( \mathbf{x}, \tilde{\mathbf{x}}_i \right ) =3/4$, the number of local maxima differs  widely. We note, however, that the density of local maxima decreases with increasing $N$ provided the local fitness correlation is kept  fixed.

\section{Model}\label{sec:model}

\begin{figure}[!h]
\centering\includegraphics[width=0.9\linewidth]{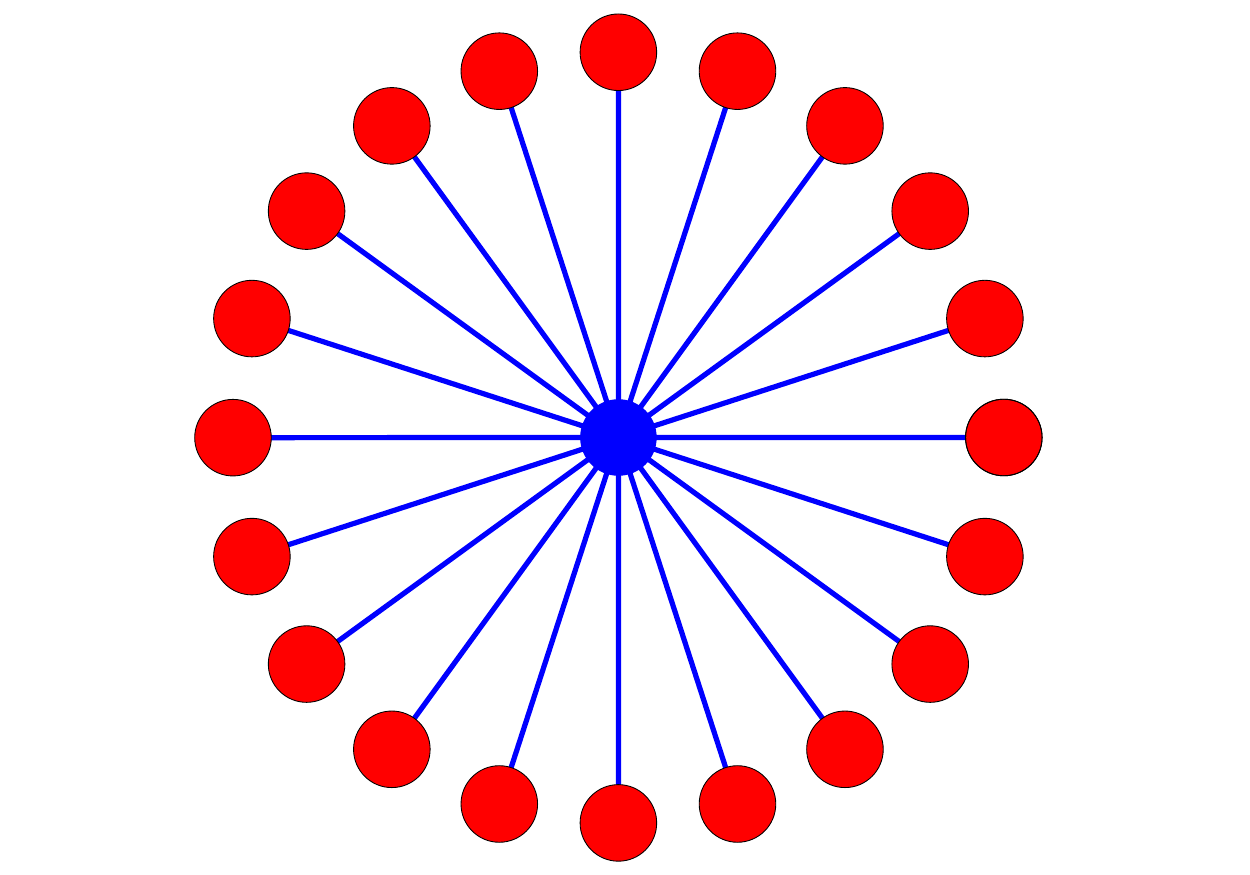}
\caption{(Color online) Star network topology  composed of $L$ peripheral operative agents that  interact with a central agent  -- the monitor. The peripheral agents search the state space and the monitor displays the fittest string at a given time. The figure illustrates the topology for $L=20$. }
\label{fig:1}
\end{figure}

Once the task   is specified we can decide on the best representation for the operative agents that will explore the state space of the problem. Clearly,
an appropriate representation for searching NK landscapes is to portray those agents as binary strings, and so henceforth  we will use the terms agent and string interchangeably. The agents are organized on a star topology  and  can interact  only with a central agent -- the monitor -- that does not search the state space but simply surveys and displays the best
performing string at  a given  moment. Figure \ref{fig:1} illustrates the topology of the communication network used in  the computational
experiments.

The $L$ peripheral strings are initialized
randomly with equal probability for the bits $0$ and $1$ and the central node displays the fittest string produced in this random setup. The search begins with the selection of one of the peripheral agents at random -- the target  agent. This  agent  can choose  between two distinct processes to move on the state space. 

The first process, which  happens with probability $p$,  is the copy  of a single bit of  the string displayed by the monitor. The copy procedure is implemented as follows.  First, the monitor string and the target string   are compared  and the different bits are singled out.  Then  one of the distinct bits  of the target string is selected at random and flipped, so  this bit is now the same in both strings.  The  second process, which happens with probability $1- p$, is the elementary move in the state space, which consists of picking a  bit  at random   from  the  target string and flipping it. This elementary move allows the agents to explore in an incremental way the $2^N$-dimensional state space. In the case the  target string is identical to the monitor string (i.e., the fittest string in the network at that time), the target   agent  flips a randomly chosen bit with probability one. 

After the target agent  is updated, we increment the time $t$ by the quantity $\Delta t = 1/L$.   Since a string operation  always results in a change of fitness,  we need to recalculate the best string and,  in case of change, update the display of the central node. Then another target agent is selected at random and the procedure described above is repeated. Note that during the increment from $t$ to $t+1$ exactly  $L$, not necessarily distinct, peripheral strings are updated.

The  search ends when one of the agents hits the global maximum and we denote by $t^*$ the halting time. The efficiency of the  search is measured by the total number of peripheral string updates necessary  to find that maximum, i.e., $Lt^*$  \cite{Clearwater_91,Clearwater_92}
and so the  computational cost of a search is  defined as $C \equiv L t^*/2^N$, where for convenience we have rescaled $t^*$ by the size of the state space $2^N$.

The parameter $p$ measures the frequency or strength of the feedback from  the monitor (inner eye) to the  peripheral operative agents. We note that the peripheral agents are not programmed to solve any task: they just flip bits at random and  occasionally copy a bit from the string displayed in the central node.  Only the central agent is capable to evaluate the goodness of the solutions.  But it is not allowed   to search the state space itself; its role is simply to evaluate and display the solutions found by the peripheral agents.  This approach is akin to the Actor-Critic model of reinforcement learning \cite{Barto_95}, in which one part of the program -- the Actor -- chooses the action to perform and the other part -- the Critic -- indicates how good this action was. 
The case $p= 0$ corresponds to the baseline situation in which    the peripheral agents do not receive any feedback from the central agent, which, however, still  evaluates  the goodness of the solutions and halts  the search when the global maximum is found.

Our model  may be viewed as a simple  reinterpretation of the well-studied model of distributed cooperative problem-solving systems based on  imitative learning \cite{Fontanari_14,Fontanari_15,Fontanari_16}. In fact,  the scenario presented above is identical to the situation where there is no central agent but each peripheral agent is linked to all others and can imitate the best agent in the network  with  probability $p$.  In that imitative learning scenario,  any agent is able to search the state space and   evaluate the quality of its solution as well as those of the other agents in the network.  The advantage of the present interpretation is that  only one special agent is endowed with the ability to evaluate the quality of the solutions, which is  clearly a very sophisticated process that  should be kept separated from the more mechanical 
state space search. Following the social brain reasoning line, the organisms   have probably  first evolved variants of this evaluative process to access their external environment, which includes the other organisms, and then modified those processes for internal evaluation.  In the present interpretation, the system exhibited in Fig.\ \ref{fig:1} is a module of the cognitive system of a single organism, whereas in the imitative learning scenario each  agent is seen as an independent organism.

\section{Results}\label{sec:res}

As a measure of  the performance of the system in searching for the global maximum of the NK landscapes, we consider the mean  computational cost $\langle C \rangle $, which  is obtained by averaging the computational cost over $10^5$ distinct searches for each  landscape realization, and the result is then averaged over  100  landscape realizations. In addition to this performance measure, we carry out diverse measurements to  get insight on the  diversity of the  strings at the halting time $t^*$.  In particular,  defining the normalized Hamming distance between the bit strings $\mathbf{x}^\alpha$ and $\mathbf{x}^\beta$ as
\begin{equation}
d \left ( \mathbf{x}^\alpha, \mathbf{x}^\beta \right ) = \frac{1}{2} -  \frac{1}{2N} \sum_{i=1}^N \left ( 1 - 2x_i^\alpha \right ) \left ( 1 - 2x_i^\beta \right ),
\end{equation}
we can introduce  the mean pairwise distance between the $L$ strings in the system,
\begin{equation}
\bar{d}= \frac{2}{L\left ( L-1 \right)} \sum_{\alpha=1}^{L-1} \sum_ {\beta = \alpha + 1}^L d \left ( \mathbf{x}^\alpha, \mathbf{x}^\beta  \right ) .
\end{equation}
This distance can be interpreted as follows:  if  we pick two strings at random, they will differ  by $N \bar{d}$ bits on average.  Hence $\bar{d}$ yields a measure of the dispersion of the strings in the state space.
The distance $ \bar{d}$  must also be averaged over the independent  searches and landscape realizations, resulting in the measure $\langle \bar{d} \rangle $.

\begin{figure}[!ht]
\centering\includegraphics[width=0.9\linewidth]{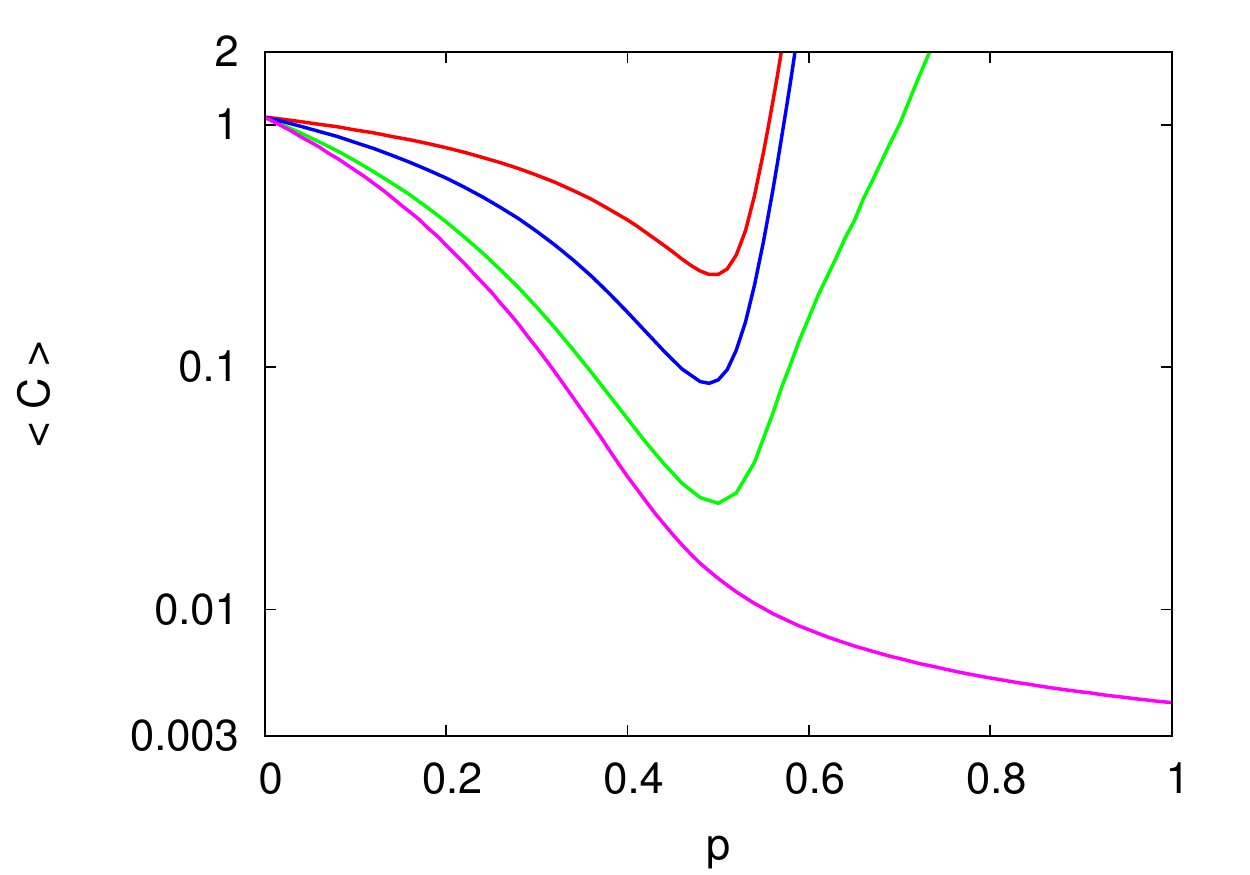}
\caption{(Color online) Mean computational cost $\langle C \rangle $  as function of the strength $p$  of the feedback between the monitor and the operative agents   for the system size $L=20$ and  (bottom to top) $K=0, 1, 3 $ and $5$. The length of the bit strings is $N=16$. Note the logarithmic scale of the axis of ordinates.
 }
\label{fig:2}
\end{figure}

We note that many applications of social heuristics to solve combinatorial problems (see, e.g., \cite{Clearwater_91,Clearwater_92,Kennedy_98,Fontanari_10}) resort to circuitous representations for  the agents as well as for their moves on the state space,  making it  difficult to gauge the complexity,  or lack thereof, of  the tasks solved by  those heuristics.  The advantage of using NK landscapes is  that we can control the difficulty of the tasks and, accordingly, in Fig.\ \ref{fig:2} we show the mean computational cost for tasks of different complexities. 

In the case of  single-maximum landscapes  ($K=0$), copying the fittest string displayed by the central node is an optimal strategy since it guarantees, on the average, a move towards the maximum. This is the reason that for a fixed system size $L$ the best performance is achieved for $p=1$.  However, the   regime of small $p$ is probably the more  relevant   since  one expects that the  feedback between the monitor and the operative agents   should  happen much less frequently than the motion in the state space.  

The presence  of local maxima ($K>0$) makes  copying  the  central node string a risky strategy since that string may display misleading information about the location of the
global maximum. In fact, the disastrous performance observed for large $p$  is  caused by the  trapping in the local  maxima,  from which escape can be extremely   costly. The culprit of the  bad performance is a groupthink-like phenomenon,  which occurs when people put unlimited faith in a  leader and so everyone in the group starts thinking alike \cite{Janis_82}.  Interestingly, the results of Fig.\ \ref{fig:2} shows that for $K>0$  there is a value  $p= p_{opt}$ that minimizes the computational cost and is practically unaffected by the complexity of the task.
However, as we will see in the following,  $p_{opt}$ decreases with increasing $L$ and increases with increasing $N$.

\begin{figure}[!ht]
\centering\includegraphics[width=0.9\linewidth]{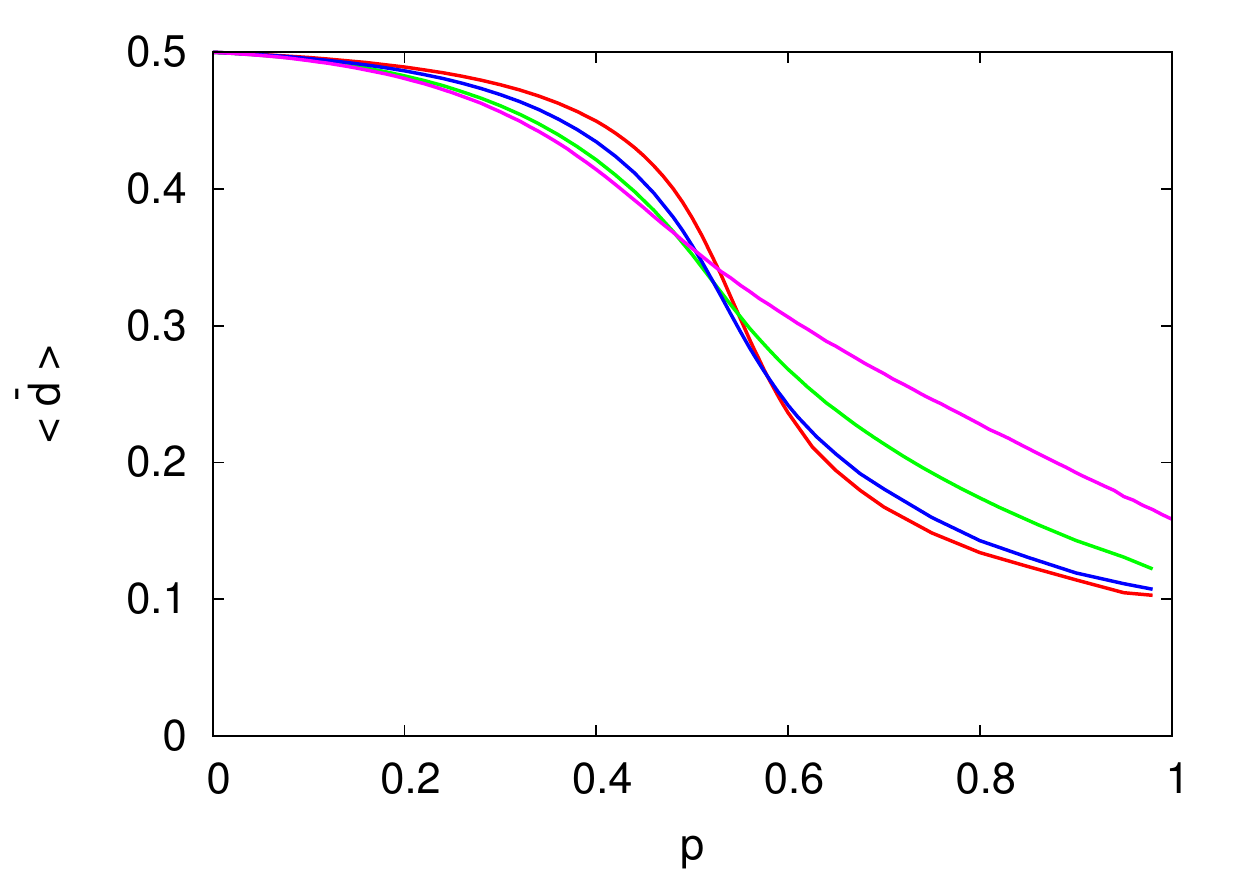}
\caption{(Color online) Mean pairwise Hamming distance $\langle \bar{d} \rangle $ measured when the search halts as function of the feedback strength $p$   for the system size $L=20$  and  (bottom to top at $p=0.4$) $K=0, 1, 3 $ and $5$. The length of the bit strings is $N=16$.
 }
\label{fig:3}
\end{figure}

Figure \ref{fig:3} offers a view of the distribution of strings in the state space at  the moment $t=t^*$ that  the global maximum is found. For a fixed task complexity (i.e., for a fixed $K$),  the mean pairwise Hamming distance  $\langle \bar{d} \rangle$  is a monotonically decreasing function of $p$, so that the strings become more similar to each other as $p$ increases, as expected. In addition, for $K > 0$ this function has an inflection point at  $p \approx p_{opt}$. Somewhat surprisingly, these results show that for  $p < p_{opt}$   the spreading of the strings  in the state space  is greater in the case of  difficult  tasks, which is clearly a good strategy to circumvent the local maxima. This behavior is reversed  in the region where the computational cost is extremely high, indicating that a large number of strings are close to the local maxima when the global maximum is found. We know that because we have  measured also the mean Hamming distance to the global maximum and 
found that this distance is greater than the typical distance between two strings.

\begin{figure}[!ht]
\centering\includegraphics[width=0.9\linewidth]{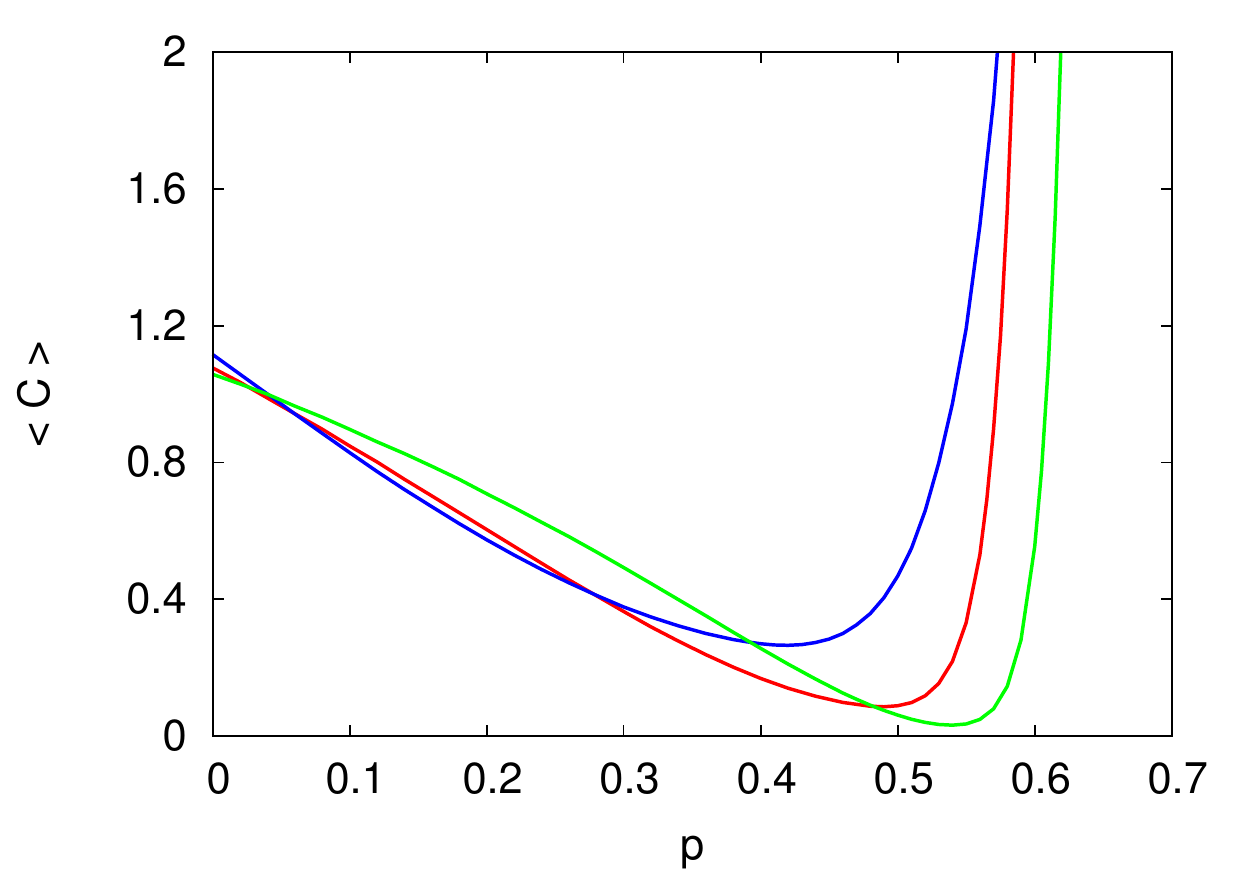}
\caption{(Color online) Mean computational cost $\langle C \rangle $  as function of the strength $p$  of the feedback between the monitor and the operative agents for  three  families of  NK landscapes (left to right at $\langle C \rangle = 1.2$): $(N=12,K=2)$, $(N=16,K=3)$ and $(N=20,K=4)$ which exhibit the same mean  local fitness correlation. The mean density of local maxima is $0.0032$, $0.0013$ and $0.0006$, respectively. The system size  is $L=20$. 
 }
\label{fig:4}
\end{figure}

To look at the effect of the state space size $N$ on the performance of the system it is convenient to vary $K$ as well so as to keep the local fitness correlation of the landscapes unchanged.  Figure \ref{fig:4} shows the mean computational cost for three families of NK landscapes with local fitness correlation equal to $3/4$ for a fixed system size. Since variation of $K$ does not affect the value of the optimal feedback strength (see Fig.\ \ref{fig:2}), the change of $p_{opt}$  observed in the figure is due to  the variation of the parameter $N$. The finding that $p_{opt}$, as well as the quality of the optimal computational  cost, increases with  the size of the problem space  indicates that the trapping effect of the local maxima is due to the density of those maxima and not to their absolute number (see Table  \ref{table:1}). We note that the  case $p=0$ can be solved analytically  (see \cite{Fontanari_15}) and  the reason that for fixed $L \ll 2^N$ the computational cost decreases with $N$ is that the chance of reverting spin flips (and hence wasting moves)  decreases as the length of the strings increases.  Only in the limit $N \to \infty$ the probability of reverting flips is zero, so that $\langle C \rangle = 1$ in that limit. 

\begin{figure}[!ht]
\centering\includegraphics[width=0.9\linewidth]{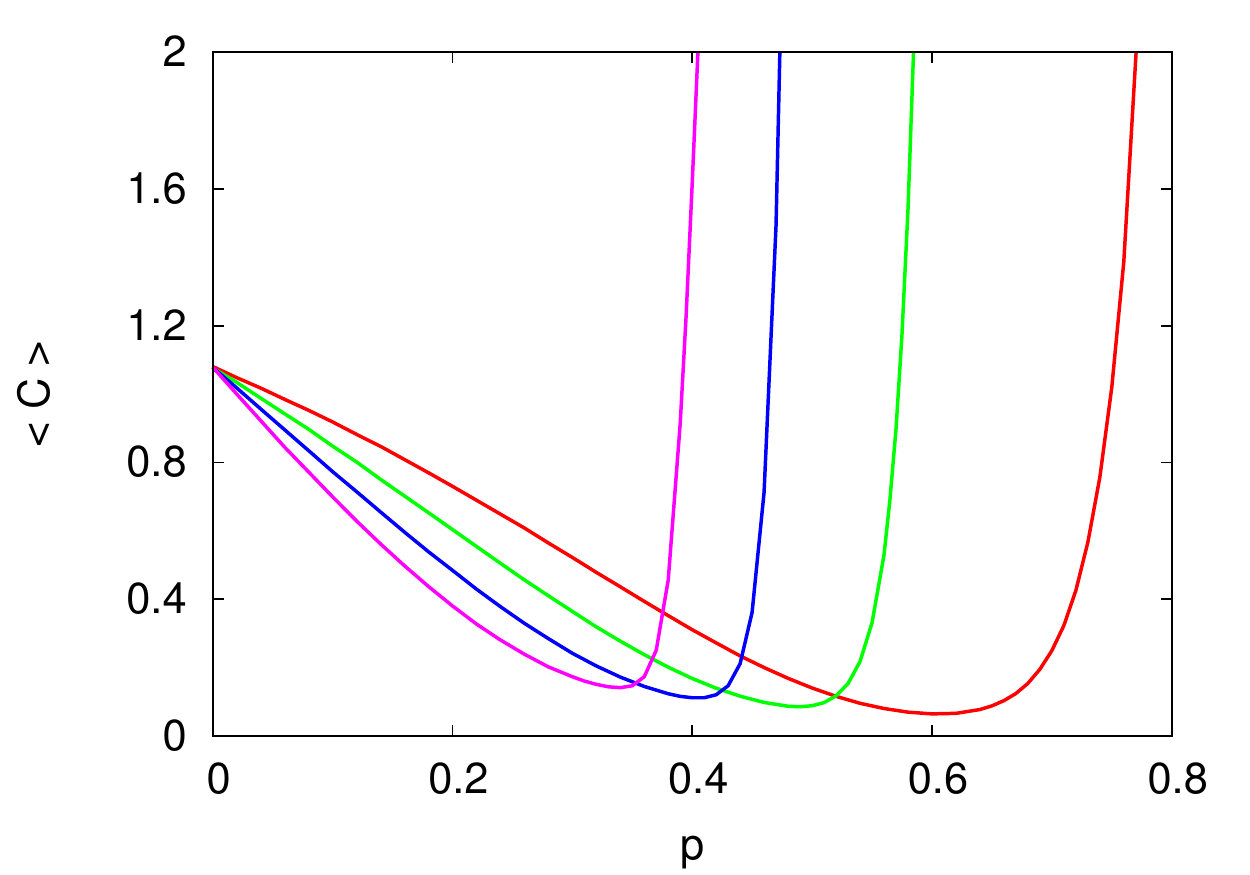}
\caption{(Color online) Mean computational cost $\langle C \rangle $  as function of the strength $p$  of the feedback between the monitor and the operative agents  
for different system sizes (top to bottom at $p = 0.2$) $L=10, 20, 40$ and $80$. The parameters of the  rugged NK landscape are $N=16$ and $K=3$.
 }
\label{fig:5}
\end{figure}

In order to offer the reader a complete view of the behavior of the system, in Fig.\ \ref{fig:5} we show the computational cost for different system sizes $L$. The results show that the optimal feedback $p_{opt}$ decreases with increasing $L$ but the quality of the optimal cost is not very sensitive to the system size. This figure reveals  also the nontrivial interplay between the system size $L$ and the feedback strength $p$. In fact,  for each $p$ there is an optimal system size that minimizes the computational cost \cite{Fontanari_15}. This optimal size decreases from infinity for $p \to 0$ to $L=2$ for $p=1$.

Finally, we note that since finding the global maxima of NK landscapes  with $K>0$ is an NP-Complete problem \cite{Solow_00}, one should not expect that the imitative  search (or any other search strategy, for that matter)   would find those maxima much more rapidly
than the independent search for a large sample of landscape realizations as that considered  here.

\section{Discussion}\label{sec:disc}

Theories of consciousness are typically expressed verbally and stated in somewhat vague and general terms even by the standards of  philosophical theories. For instance, many notorious thought experiments of the field (e.g., Mary's room \cite{Jackson_82} and the philosopher's zombie \cite{Chalmers_96})  have multiple interpretations because their specifications are unclear \cite{Dennett_91} and even the so-called `hard problem' of consciousness  (i.e., how physical processes in the brain give rise to subjective experience \cite{Chalmers_96}) is viewed by some researchers as a hornswoggle problem \cite{Churchland_96}  and a major misdirection of attention \cite{Dennett_96}.

Perhaps what is missing is an effort to express  theories of consciousness, or at least some of their premises, as computer programs 
\cite{Perlovsky_01}. This
would require a  complete and detailed specification of all assumptions, otherwise the program  would not run in the computer \cite{Cangelosi_02}. (Of course, this research program does not apply to those theories that are  built on the premise of the impossibility of such a computer simulation.)
 With very few exceptions (see, e.g., \cite{Santos_15}) computer simulations and mathematical models  have greatly aided   the elucidation of nonintuitive issues on Evolutionary Biology \cite{Dawkins_86}, and we see no intrinsic reason that could prevent the use of those tools  to verify assumptions and predictions of  theories  of  consciousness, particularly of  those theories that grant a selective value to consciousness.
  
In this paper we explore a key  element of the theories that view consciousness as a schematic description of the brain's states, namely, the existence of  inner-eye-like processes that monitor those states and provide feedback on their suitability to the attainment of the organism's goals \cite{Humphrey_99,Graziano_13}. We use a cartoonish model of this scenario, in which a group of operative agents search blindly for the  global maximum of a fitness landscape and a monitor provides them with feedback on the quality (fitness) of the proposed
solutions. This feedback is then used by the operative agents to bias their searches towards the (hopefully) fittest regions of the landscape. We
interpret this self-monitoring  as the awareness of the system about the computation it is carrying out.

We find that a weak feedback between the monitor and the operative agents improves the performance of the system, regardless of the difficulty of the task, which is gauged by the number of local maxima in the landscape. In the case of easy tasks (i.e., landscapes without local maxima), the performance improves monotonically as the feedback strength  increases, but for difficult tasks  too much feedback leads to a disastrous performance (see Fig.\ \ref{fig:2}). Of course, one expects that the value of the feedback strength, which measures the influence of the inner-eye process on the low-level cognitive processes,  will be determined by  natural selection  and so it is likely to be  set to an optimal value that guarantees the maximization of the system performance.

In closing, our findings suggest that the inner-monitor\-ing of the system behavior (computations, in our case), which is a key element in some theories of consciousness \cite{Humphrey_99,Graziano_13}, results in an improved general problem-solving capacity. However, if a system that, in the words of Dennett \cite{Dennett_91},
  ``... monitors its own activities, including even its own internal activities, in an indefinite upward spiral of reflexivity''  can be said to be conscious is an issue that is best left to the philosophers,

\acknowledgments
The research of JFF was  supported in part by grant
15/21689-2, S\~ao Paulo Research Foundation
(FAPESP) and by grant 303979/2013-5, Conselho Nacional de Desenvolvimento 
Cient\'{\i}\-fi\-co e Tecnol\'ogico (CNPq).

\end{document}